\documentclass{ecai}
\usepackage{times}
\usepackage{graphicx}
\usepackage{latexsym}
\usepackage{amsmath}
\usepackage{amsmath, bm}
\usepackage{amssymb}
\usepackage{booktabs}
\usepackage{multirow}
\usepackage{epstopdf}
\usepackage{url}
\usepackage[colorlinks,linkcolor=blue,citecolor=blue,urlcolor=blue,backref=page]{hyperref}
\usepackage{breakurl}
\ecaisubmission   % inserts page numbers. Use only for submission of paper.
\begin{document}

\title{Predicting Long-Term Skeletal Motions by a Spatio-Temporal Hierarchical Recurrent Network}
% generating recognizable long-term articulated objects motion via hierarchical recurrent network

\author{Junfeng Hu \institute{School of Big Data \& Software Engineering, Chongqing University, China,
IEEE student member,
email: hjf@cqu.edu.cn}
\and Zhencheng Fan \institute{School of Big Data \& Software Engineering, Chongqing University, China,
email: zanecode6574@cqu.edu.cn}
\and Jun Liao\institute{School of Big Data \& Software Engineering, Chongqing University, China,
email: liaojun@cqu.edu.cn}
\and Li Liu\institute{School of Big Data \& Software Engineering, Chongqing University, China,
email: dcsliuli@cqu.edu.cn; corresponding author}}

\maketitle
\bibliographystyle{ecai}

\begin{abstract}
The primary goal of skeletal motion prediction is to generate future motion by observing a sequence of 3D skeletons. A key challenge in motion prediction is the fact that a motion can often be performed in several different ways, with each consisting of its own configuration of poses and their spatio-temporal dependencies, and as a result, the predicted poses often converge to the motionless poses or non-human like motions in long-term prediction.
This leads us to define a hierarchical recurrent network model that explicitly characterizes these internal configurations of poses and their local and global spatio-temporal dependencies. The model introduces a latent vector variable from the Lie algebra to represent spatial and temporal relations simultaneously. Furthermore, a structured stack LSTM-based decoder is devised to decode the predicted poses with a new loss function defined to estimate the quantized weight of each body part in a pose.
Empirical evaluations on benchmark datasets suggest our approach significantly outperforms the state-of-the-art methods on both short-term and long-term motion prediction.
\end{abstract}

\section{Introduction}
\label{sec:sec2}
Human or animals motion prediction has become an important research field, given its role in facilitating a broad range of applications in sports, healthcare, education, security, virtual and augmented reality, among others. Current techniques are becoming mature to predict short-term motions from 3D skeleton collection devices like Kinect. For example, motions like \emph{hand posing}, which contains a series of hand poses, can be collected by depth sensors with their 3D skeleton data recording the trajectories of human body joints. Each pose consists of a fixed number of bones and joints and can be inferred from a video frame (in other words, each frame only contains one pose).
The main focus of this paper is on long-term motions, where a motion is a collection of spatio-temporally related poses.
\begin{figure}
\centering
\includegraphics[width=8.5cm]{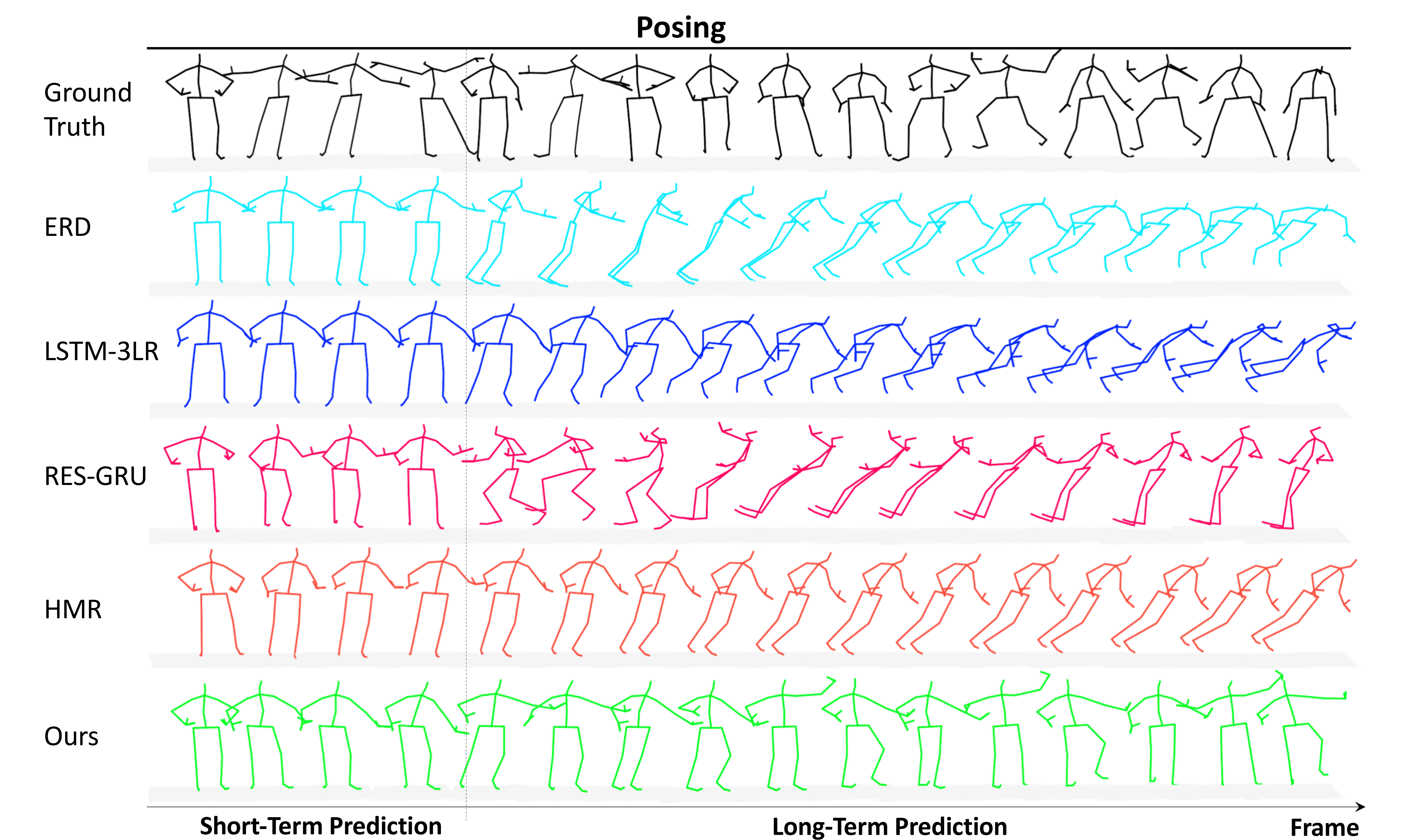}
\caption{Visualization of motion predictions (i.e. hand posing) by different models compared with ground truth. The first four frames are short-term predicted poses and the rest are long-term predicted poses. }
\label{fig:fig4}
\end{figure}
As illustrated in Fig~\ref{fig:fig4}, given a sequence of ground truth frames, many existing models, such as Encoder-Recurrent-Decoder network (ERD)~\cite{Fragkiadaki2015Recurrent}, LSTM 3 layers (LSTM-3LR)~\cite{Fragkiadaki2015Recurrent}, Residual Gated Recurrent Unit (RES-GRU)~\cite{Martinez2017On} and Hierarchical Motion Recurrent Network (HMR)~\cite{Zhenguang2019Towards}, can generate short-term motions, which are normally regard as less than 400 milliseconds (e.g. the first four frames in this example). However, the poses become unrecognizable or motionless on long-term prediction.
It is well known that modeling motions naturally requires the characterization of their spatio-temporal dependencies among poses. That is to say, a long-term motion prediction model should capture inherent structures associated with individual poses as well as their spatio-temporal dependencies.

Despite being a very challenging problem, in recent years there has been a rapid growth of interest in modeling and predicting articulated object motions. Conventional approaches have gained attention in recent years for addressing object motion prediction problems. They leverage expert knowledge about kinematics and utilize latent-variable models like hidden Markov assumptions~\cite{Lehrmann2014,Lehrmann2013,Vladimir2001Learning}, Gaussian process~\cite{Wang2007Gaussian}, Boltzmann machine~\cite{Taylor2006Modeling} and implicit probabilistic model~\cite{Sidenbladh2002Implicit} to characterize motion sequences. But motions and their spatio-temporal relations in these models need to be manually encoded, which could be rather difficult to scale up and is almost impossible for many practical scenarios where spatio-temporal relations among poses are intricate.

On the other hand, the most popular modeling paradigm might be that of the deep neural networks, which include techniques such as recurrent neural network (RNN), long short-term memory models (LSTM) and gated recurrent unit (GRU). While these neural network-based approaches are capable of managing temporal contexts, they have difficulties in capturing long-term dependencies~\cite{Zhenguang2019Towards}. This is because these models rely on conventional recurrent units where the hidden state sequentially reads a frame and updates its value, which leads to overwhelming state estimation from the inputs in recent time steps.
In particular, they suffer from \emph{first frame discontinuity}, that is, a prominent jump between the last ground truth frame and the first predicted pose. In addition, current works mainly focus on temporal information and are unfortunately rather limited in characterizing rich fine-grained spatial relationships among joints. In fact, as these models mostly focus on coarse-grained (high-level) spatial information  (e.g. taking all the joints as a whole in a pose), ignoring internal joints dependency, only spatial relations associated with entire body can be sufficiently captured. As a result, the predicted human poses often converge to the mean (i.e. motionless) poses~\cite{Xue2016Visual} or shift to unrecognizable (e.g. non-human like) motions~\cite{Martinez2017On} in long-term prediction, as illustrated in Fig~\ref{fig:fig4} (e.g. ERD, LSTM-3LR, RES-GRU and HMR). Moreover, most of the existing approaches adopt \emph{walking} activity, which only repeats a fixed style of regular movements of legs, to demonstrate their superiority on long-term prediction. However, we found these models can only perform well on such simple activities but not others (e.g. \emph{eating} and \emph{posing}), especially for the activities without any explicit discipline.

\begin{figure}
\centering
\includegraphics[width=6cm]{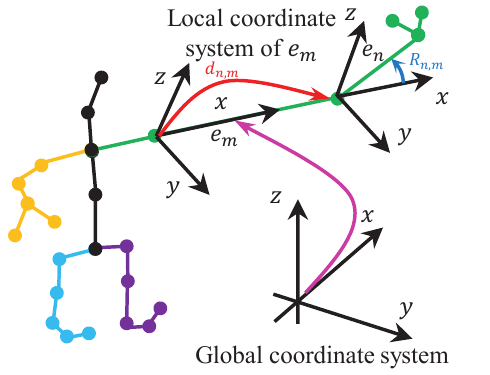}%{Fig6.eps}
\caption{Illustration of skeletal rotation ($R_{n,m}$) and translation ($d_{n,m}$) on Lie group. In our model human skeleton is divided into five kinematic chains, i.e., spine (black), right arm (yellow), left arm (green), right leg (cyan) and left leg (violet).}
\label{fig:fig6}
\end{figure}

To address these issues in long-term motion prediction, we present a spatio-temporal hierarchical recurrent neural network to explicitly model the motion context of spatio-temporal relations and predict future motions. In particular, our approach considers a principled way of dealing with the inherit structural variability in long-term motions. Briefly speaking, to describe an articulated object, we propose to introduce a set of latent vector variables generated from the Lie algebra to represent several separate kinematic chains of body part movements as shown in Fig~\ref{fig:fig6}.
Now each resulting vector from the Lie algebra-based representation contains its unique set of poses that together with the corresponding spatial features and temporal information. To fully characterize a certain cluster of instances that possess similar motions and their spatio-temporal dependencies, a hierarchical recurrent network is devised to encode the spatial relationships along with the temporal relations. Specifically, in each recurrent layer, a unit variable that represents a bone in a frame is updated by exchanging information with other unit variables considering both spatial and temporal dependencies. Also, a global spatial state and a global temporal state are incorporated into the unit hierarchically in each layer to capture global spatio-temporal relations. Different from traditional recurrent units, such as LSTM and GRU, all the units in our network can hierarchically read unit states from the previous step and update their values simultaneously within one current step. In this way, spatio-temporal information can be maintained in each recurrent step, allowing our model to capture long-term dependencies.
In addition, a structured stack LSTM-based decoder is introduced to decode the predicted poses with a new loss function defined to estimate the importance of a bone quantitatively concerning its kinematic location and length in the skeleton. In this way, our neural network-based model is more capable of characterizing the inherit structural variability in long-term motion prediction when comparing to existing methods, which is also verified during empirical evaluations to be detailed in later sections.
Our project's main page with experimental videos and the official code is available at \url{https://github.com/p0werHu/articulated-objects-motion-prediction}.

\section{Related work}
In this section, a brief review of related topics, i.e. motion representation, modeling and prediction, are listed below.
\subsection{Pose and motion representation}
As the fundamental issue in motion related applications, three vision-based approaches are commonly used to represent poses, namely, RGB-based representation~\cite{45441,Lotter2016Deep,Mathieu2015Deep}, depth map-based representation~\cite{Li2016Recognizing} and skeleton-based representation. Here we mainly focus on the skeleton-based representation. Currently, skeleton-based representation has attracted large attention because of its immunity to viewpoint change~\cite{Han2017Space} and the geometric description of rigid body~\cite{Murray:1994:MIR:561828}. The existing approaches are roughly divided into two categories: joint-based approaches~\cite{Liu2018Skeleton,Butepage2017Deep,Tang2018Long,Du2015Hierarchical} which regard skeleton as a set of independent points and part-based (or bone-based) approaches~\cite{Grassia1998Practical,Jain2016Structural,Martinez2017On} which consider skeleton as a set of rigid segment made up of two joint points~\cite{Vemulapalli2014Human}.
%For instance, joint-based approaches~\cite{Liu2018Skeleton,Butepage2017Deep,Tang2018Long,Du2015Hierarchical} feeded raw 3D joint positions to the model as the human pose representation, and part-based approaches~\cite{Grassia1998Practical,Jain2016Structural,julieta2017motion} describe the orientation and angle between any two bones using exponential map.
%Du et. al~\cite{} used raw 3D joint locations but divided the human skeleton into five different parts and feed them into separate sub-networks. A tree structure-based skeleton traversal approach~\cite{Liu2018Skeleton} was proposed to probe shallow kinematic relationship between joints, but it may lead to high dimension problem, which could be rather difficult to apply in prediction tasks.

Besides, motion representation is also very significant that it should effectively capture the spatial motion characteristics of joints or bones. Two most common methods are Euler angle representation and unit quaternion representation. However, the Euler angle representation suffers from non-intrinsic singularity or gimbal lock issue, which leads to numerical and analytical difficulty, while the unit quaternions approach leads to singularity-free parametrization of rotation matrices, but at the cost of one additional parameter~\cite{JonghoonGeometric}.
Currently, Lie group-based representation was proposed to solve these singularity and computational issues in manifold-based skeletal motions. Vemulapalli et al.~\cite{Vemulapalli2014Human} first introduced a Lie group, named Special Euclidean group SE(3), in skeletal motion representation to calculate the relative geometry between various body parts. It is found that the relative geometry provides a more sensible description compared to absolute locations of one bone over SE(3) representation. On the other hand, Special Orthogonal Group SO(3)~\cite{Vemulapalli2016Rolling,Huang2017Deep}, another Lie group, was utilized to represent only rotations but not translations in motions, which obtained similar performance as SE(3). However, all the joints in these approaches are regarded equally in a skeleton by ignoring the anatomical restricts among chains~\cite{Zhenguang2019Towards}. This inspires us to divide an articulated object into several kinematics chains to retain these skeletal restrictions.

\subsection{Motion modeling}
Motion prediction requires a model having an efficient encoding capability on input motion sequences. Initiatively linear SVM were adopted to model human motion~\cite{Vemulapalli2014Human,Ohn2013Joint}, with Lie group skeleton representation to characterize the spatial and temporal features~\cite{Vemulapalli2016Rolling}. Lv et al.~\cite{Lv2006Recognition} leveraged Hidden Markov Models (HMMs) to capture the sequential properties of poses. Recently RNNs become the most popular model. Du et al.~\cite{Du2015Hierarchical} used RNN by dividing the human skeleton into five parts and learning their features separately, which are integrated by a single layer network afterwards.
Huang et al.~\cite{Huang2017Deep} incorporated the Lie group into a recurrent network structure enabling it the ability to learn more appropriate spatio-temporal features than SVM and HMM. Similar to CNNs, this network defines RotMap layer as convolutional layer and RotPoling layer as pooling layer. Liu et al.~\cite{Liu2018Skeleton} proposed a spatio-temporal LSTM for motion modeling with a novel trust gate introduced to reduce noise caused by data collection devices.
However, the motivation of these models mainly focused on designing efficient encoders, which refine high-level encoding features, but many significant spatio-temporal dependencies are neglected. Consequently, these encoders cannot be transplanted to motion prediction directly.
Different from the previous work, we design a RNN-based encoder to capture the spatio-temporal features of input pose sequences in one single step and use hierarchical structures to retain long-term spatio-temporal information.

\subsection{Motion prediction}
As aforementioned in the introduction section, many conventional approaches need to handcraft spatio-temporal relations and even their weights from domain knowledge in motion prediction. Therefore, deep neural networks are commonly used to predict future motions in recent years. Fragkiadaki et al.~\cite{Fragkiadaki2015Recurrent} proposed the ERD model that incorporates nonlinear encoder and decoder networks before and after recurrent layers and a LSTM in the recurrent layer. SRNN~\cite{Jain2016Structural} divides human body into three different parts (i.e. spine, arms, and legs) among which the spatial and temporal relations are learnt separately. RES-GRU~\cite{Martinez2017On} is a sequence-to-sequence architecture that combines GRU and residual connection in the decoder. HMR~\cite{Zhenguang2019Towards} introduces a hierarchical motion recurrent network, which exchanges information with neighboring frames to obtain temporal features of motions. Li et al.~\cite{Li2018Convolutional} proposed a hierarchical structure of CNN to model human dynamics.
Tang et al.~\cite{Tang2018Long} proposed a modified highway unit (MHU) and a gram matrix loss function for long-term prediction, attempting to reduce the problem of motionless.
To address the problems in these models (as mentioned in the introduction section), we present the hierarchical recurrent network model to explicitly capture the inherent structural varieties of skeleton motions with spatio-temporal dependencies.

\section{Lie algebra representation for skeletal data}

It is known that the relative geometry of a pair of two body parts of one skeleton can be described by representing each of them in a local coordinate system attached to the other~\cite{Vemulapalli2014Human}. Given two bones $e_m$ and $e_n$, as shown in Fig~\ref{fig:fig6}, the local coordinate system of $e_m$ is computed by rotating with minimum rotation and translating the global coordinate system so that $e_m$ becomes the position and orientation of $x$-axis (i.e. its starting joints becomes the origin and the $x$-axis is aligned with it). After this process, we can obtain the location of $e_n$ attached to the local system of $e_m$, denoted by $e^m_n$. Then, we can compute a 3D rigid transformation formalized as \begin{footnotesize}$\begin{pmatrix}  R_{n,m} & d_{n,m} \\ 0 & 1  \end{pmatrix}$, where $R_{n,m}$\end{footnotesize} is a $3\times 3$ rotation matrix and $d_{n,m}$ is a 3D translation vector to take $e_m$ to the position and orientation of $e_n$.
\begin{equation}
\footnotesize
\begin{bmatrix} e^m_{n,end} \\ 0 \end{bmatrix} = \begin{bmatrix} R_{n,m} & d_{n,m} \\ 0 & 1 \end{bmatrix} \begin{bmatrix} \ell_n \\ 0 \\ 0 \\ 1 \end{bmatrix},
\end{equation}
where $e^m_{n,end}$ means the end joint of $e^m_n$ and $\ell_n$ means the length of $e_n$. Similarly, the location of $e_m$ attached to the local system of $e_n$ is calculated by another transformation matrix. As a result, a total number of $M\times (M-1)$ transformation matrices are obtained where $M$ is the number of bones. Mathematically, 3D rigid transformation is element of the Special Euclidean group SE(3). In the end, one skeleton is represented as a curve in $SE(3)\times ...\times SE(3)$.

However, similar to the process in ~\cite{Vemulapalli2014Human,Vemulapalli2016Rolling,Zhenguang2019Towards}, we fix the bone length by a normalized bone length, indicating that all the translation vectors are static; and thus, only the rotation matrix is required in our model, which is different from the unnecessary representation using SE(3) in ~\cite{Vemulapalli2014Human,Vemulapalli2016Rolling,Zhenguang2019Towards}. Meanwhile, given that a human body is described by a kinematic tree consisting of five kinematic chains (i.e. spine, two legs, and two arms), as illustrated in Fig~\ref{fig:fig6}, we only need to calculate rotation matrix between two neighbouring bones sharing the same joint instead of two arbitrary bones within one chain.
In this way, the structure of skeletal anatomy is maintained in terms of the anatomical restricts among chains. In addition, the number of rotation matrices in our model is reduced, which may potentially decrease computational cost compared with those containing any pair of bones~\cite{Vemulapalli2014Human,Vemulapalli2016Rolling}.
In practice, we first compute the axis-angle representation ($\textbf{n}$, $\theta$) by
\begin{equation}
\footnotesize
\textbf{n} = \frac{corss(e_n, e_m)}{||corss(e_n, e_m)||},
\end{equation}
\begin{equation}
\footnotesize
\theta = \arccos(e_n\cdot e_m),
\end{equation}
where $cross$ denotes outer and $\cdot$ means inner products. Then, the rotation matrix $R_{n,m}$ is calculated by Rodriguez formula:
\begin{equation}
\footnotesize
R_{n,m} = I + \sin(\theta)\textbf{n}^\land + (1-\cos(\theta))\textbf{n}^{\land2},
\end{equation}
where $I\in\mathbb{R}^{3\times 3}$ is a identity matrix and $\textbf{n}^\land$ is the skew-symmetric matrix of $\textbf{n}$. Note that the set of rotation matrices belong to the Special Orthogonal Group $SO(3)$, the skeleton is represented as a curve in $SO(3)\times ...\times SO(3)$.

Because regression in the curved space $SO(3)\times ...\times SO(3)$ is non-trivial, we map this curved space to its tangent space regarded as Lie algebra $\mathfrak{so}(3)\times ...\times \mathfrak{so}(3)$ using the approximate solution~\cite{Huang2017Deep} of logarithm map:
\begin{equation}
\footnotesize
\omega(R_{n,m}) = \frac{1}{2\sin(\theta(R_{n,m}))} \begin{bmatrix} R_{n,m}(3,2) - R_{n,m}(2,3) \\ R_{n,m}(1,3) - R_{n,m}(3,1) \\ R_{n,m}(2,1) - R_{n,m}(1,2) \end{bmatrix},
\end{equation}
\begin{equation}
\footnotesize
\theta(R_{n,m}) = \arccos(\frac{Trace(R_{n,m}) - 1}{2}).
\end{equation}
In the end, the skeleton are mapped to a series of Lie algebra vectors: $\bm{\omega}=[{\omega^1_1}^\mathrm{ T }, ...,{\omega^1_{K_1}}^\mathrm{ T }, ..., {\omega^C_1}^\mathrm{ T }, ..., {\omega^C_{K_C}}^\mathrm{ T }]^\mathrm{ T }$, where $C$ denotes the number of chains (in our model $C=5$ for human motion) and $K_c$ ($c\in\{1,\ldots,C\}$) equals the number of bones in the $c$-th chain minus one.

\section{Our model}
\begin{figure*}
\centering
\includegraphics[width=1\textwidth]{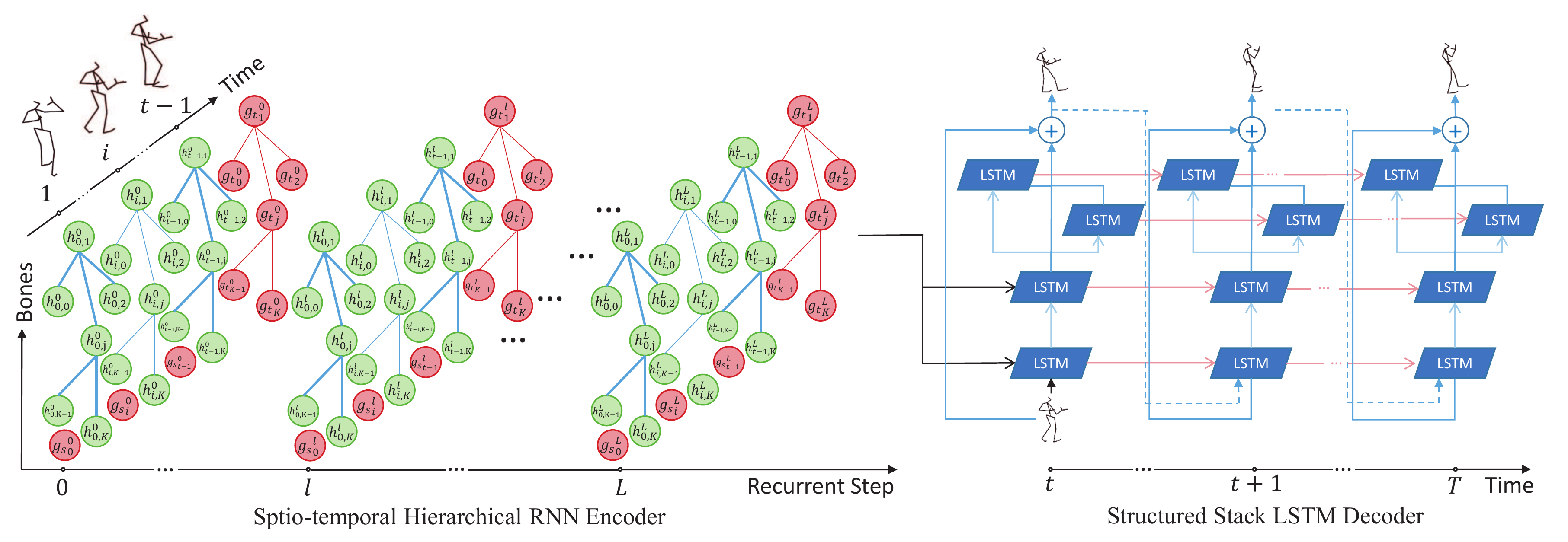}%{Fig1.eps}
\caption{The overview of our model. $h^l_{i,j}$ represents the local hidden state of bone $j$ at frame $i$ in layer $l$. ${g_t}^l_{j}$ represents global temporal state of bone $j$ in layer $l$, and ${g_s}^l_{i}$ represents global spatial state of a pose at frame $i$ in layer $l$. $L$ is the number of recurrent layers.}
\label{fig:fig1}
\end{figure*}

Given a sequence of observed poses $\boldsymbol{P} = (p_1, p_2, ..., p_t)$ in a motion, the goal is to predict its future poses $\boldsymbol{\hat{P}} = (p_{t+1}, p_{t+2}, ..., p_{T})$, where $T$ is the number of frames. %Conventional methods only represented 3D skeleton in raw European Space and ignore the anatomical restricts of skeleton, which causes the distortion of the predicted pose. Furthermore, traditional models like RNN and LSTM, pay more attention to the latter input poses so that they have difficulties to model long-term predictions~\cite{Martinez2017On}. Meanwhile, decoders of existing models of this field normally output all the prediction skeletons at one time, which violates the intuition of motion prediction. Given that the spine is the backbone of the human pose, it is more sensible to forecast the spine at first and then predict other parts according to the spine.
Our model is divided into two parts, a spatio-temporal hierarchical RNN encoder and a structured stack LSTM decoder, as shown in Fig~\ref{fig:fig1}. The encoder aims to model the motion efficiently, which encodes all the observed poses in $\boldsymbol{P}$ simultaneously concerning their sptio-temporal dependencies. At each recurrent layer, a unit on behalf of a bone in a frame exchanges information with two neighboring units at the spatial axis and the previous unit at the temporal axis. Meanwhile, global states on both temporal and spatial dependencies are incorporated into the unit to help our model maintain global information. On the other hand, the decoder is designed to predict future poses $\boldsymbol{\hat{P}}$. In the first layer, the decoder deciphers overall information from the previously encoded features. Next, a spine LSTM is first used to decode a spine pose in the second layer, and then another two LSTMs are utilized (i.e. leg LSTM and arm LSTM) to decode two arms and two legs according to the previously decoded spine, respectively. Note that the first $t-1$ frames will be feeded into the encoder, and the last frame $t$ will be used in the decoder.

\subsection{Spatio-temporal hierarchical RNN encoder}

\begin{figure}
\centering
\includegraphics[width=8.5cm]{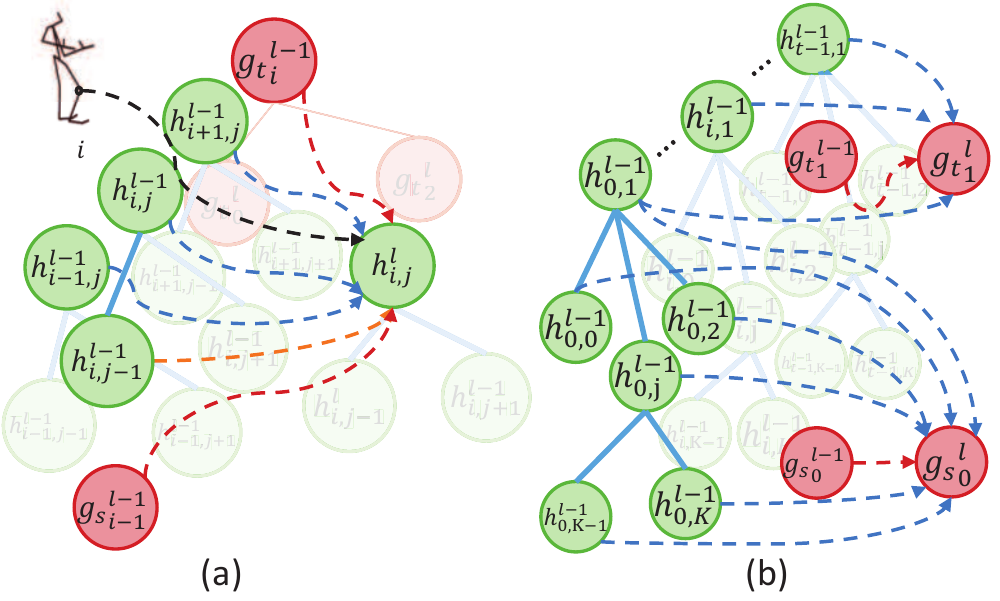}%{Fig2.eps}
\caption{Illustrations of dataflows that update local and global states in our model.
(a) $h^l_{i,j}$ is updated by exchanging information with states in spatial axis (orange line), temporal axis (blue lines) and global axis (red lines).
(b) ${g_t}_{j}^{l-1}$ and ${g_s}_{i}^{l-1}$ are updated according to their previous layer's states as well as all hidden states in their corresponding axis.}
\label{fig:fig2}
\end{figure}
We use $i\in (1, ..., T)$ and $j\in (1, ..., K)$ (where $K=\sum_{c=1}^{C}K_c$) to denote the index of a frame and an element in the Lie algebra vector $\bm{\omega}$, respectively. Although not accurate, we call an element in $\bm{\omega}$ as a \emph{bone} for convenience. As shown in Fig~\ref{fig:fig2} (a), a state $h^l_{i,j}$ in layer $l$ is updated by exchanging information with its neighbors $h^{l-1}_{i-1,j}$, $h^{l-1}_{i+1,j}$, $h^{l-1}_{i,j}$  at temporal axis and $h^{l-1}_{i,j-1}$ at spatial axis. In this way, the model learns the fined-grained features of current bone on both spatial and temporal dependencies. In details, with the increase of the recurrent steps, $h^l_{i,j}$ exchanges context with more bones $j$ in different frames and bones in the current frame $i$. Besides, global spatial state ${g_s}_{i}^{l-1}$ and temporal state ${g_t}_{j}^{l-1}$ are used to incorporate global features into the state $h^l_{i,j}$. The purpose is that ${g_s}_{i}^{l-1}$ represents global feature of bones in frame $i$ so that the model obtains the high-level information of the current pose. ${g_t}_{j}^{l-1}$ denotes global information about bone $j$ at each frame, which enables the model to encode the movement of this bone within one state. At the first recurrent layer, we initialize hidden and cell states such that $h^0_{i,j} = c^0_{i,j} = Wp_{i,j} + b$, ${g_t}_{j}^0 = {c_{g_t}}_{j}^0 = \frac{1}{t-1}\sum\limits_{i=1}^{t-1}h^0_{i,j}$, and ${g_s}_{i}^0 = {c_{g_s}}_{i}^0 = \frac{1}{K}\sum\limits_{j=1}^Kh^0_{i,j}$, where $W$ and $b$ are parameters in the network.

To update $h^l_{i,j}$, by following the design of LSTM, there are six different forget gates to control the feature flows from six incoming context channels separately (i.e. three flows from temporal axis, one flow from spatial axis, and two global flows): ${l^l}_{i,j}$, ${f^l}_{i,j}$, ${r^l}_{i,j}$, ${s^l}_{i,j}$, ${gs^l}_{i,j}$, and ${gt^l}_{i,j}$. The input gate $in_{i,j}^l$ and output gate $out_{i,j}^l$ control the information flow from input pose $p_{i,j}$ to update the hidden state $h^l_{i,j}$ of this recurrent layer.
The process of obtaining the input gate $in^l_{i,j}$ can be formulated as below:
%The process of updating cell $c^l_{i,j}$ and $h^l_{i,j}$ can be formulated as:
\begin{equation}
\footnotesize
\begin{split}
in_{i,j}^l = \sigma(U_{in}p_{i,j} + W_{in}(h^{l-1}_{i-1,j}, h^{l-1}_{i+1,j}, h^{l-1}_{i,j}) \\ + Z_{in}h^{l-1}_{i,j-1} + B_{in}{g_s}_{i}^{l-1} + G_{in}{g_t}_{j}^{l-1} + b_{in})
\end{split}
\end{equation}
Moreover, formulations of obtaining forget gates for the three spatial, two temporal, two global information flows and the output gate are similar with the formulation above except parameters $U$, $W$, $Z$, $B$, $G$, and $b$.
%As for the output gate that determines how the current hidden state $h^l_{i,j}$ is derived from the current internal cell $c^l_{i,j}$,
As for the modulated input $\tilde{c}^l_{i,j}$, we replace the sigmoid activation function (i.e. $\sigma$) by the $tanh$ activation function. Then, the process of updating memory cell $c^l_{i,j}$ and hidden state $h^l_{i,j}$ can be formulated as:
\begin{equation}
\footnotesize
\begin{split}
c^l_{i,j} = in^l_{i,j} \odot \tilde{c}^l_{i,j} + l^l_{i,j} \odot c^{l-1}_{i-1,j} + f^l_{i,j} \odot c^{l-1}_{i,j}\\
+ r^l_{i,j} \odot c^{l-1}_{i+1,j} + s^l_{i,j} \odot c^{l-1}_{i,j-1} \\+ {gs^l}_{i,j} \odot {c_{g_s}}_{i}^{l-1} + {gt}^l_{i,j} \odot {c_{g_t}}_{j}^{l-1},
\end{split}
\end{equation}
\begin{equation}
\footnotesize
h^l_{i,j} = out^l_{i,j} \odot tanh(c^l_{i,j}),
\end{equation}
where $\odot$ means Hadamard product. Note that for all $c^l_{i,j}$, the parameters in the same frame $i$ are shared within the layer $l$, and also the parameters are shared among different recurrent layers.

To update temporal global state ${g_t}_{j}^{l}$ and spital global state ${g_s}_{i}^{l}$, as shown in Fig~\ref{fig:fig2} (b), for ${g_t}_{j}^{l}$, we first design forget gates for cell $c^l_{i,j}$ of all frames $i\in (1, ..., t-1)$ and then a forget gate for ${c_{g_t}}_{j}^{l-1}$ is introduced. %This process is formulated as:
The process of getting forget gates for $c^l_{i,j}$ and ${c_{g_t}}_{j}^{l-1}$ are formulated as:
\begin{equation}
{f^l_c}_i = \sigma(\tilde{W}_c h^l_{i,j} + \tilde{Z}_c {g_t}_{j}^{l-1} + b_c)\quad i\in (1, ..., t-1),
\end{equation}
\begin{equation}
\label{equ:equ2}
f^l_{gt} = \sigma(\tilde{W}_{gt} (\frac{1}{t-1}\sum\limits^{t-1}_{i=1}h^l_{i,j}) + \tilde{Z}_{gt} \,{g_t}_{j}^{l-1} + \tilde{b}_{gt})
\end{equation}
The output gate $out^l_{gt}$ is parallel to Equation~\ref{equ:equ2} with parameters $\tilde{W}_o$, $\tilde{Z}_o$, and $\tilde{b}_o$. Then, the next two equations show the procedure of obtaining ${c_{g_t}}_{j}^{l}$ and ${g_t}_{j}^{l}$:
\begin{equation}
\footnotesize
{c_{g_t}}_{j}^{l} = \sum\limits^{t-1}_{i=1}{f^l_c}_i \odot c^l_{i,j} + f^l_{gt} \odot {c_{g_t}}_{j}^{l-1},
\end{equation}
\begin{equation}
\footnotesize
{g_t}_{j}^{l} = out^l_{gt} \odot tanh({c_{g_t}}_{j}^{l}),
\end{equation}
Similarly, for global state ${g_s}_{i}^{l}$, we design forget gates for all the cells $c^l_{i,j}$ of one frame $i$ and ${c_{g_s}}_{i}^{l-1}$. The details of omitted formulations are provided in our project home page due to limited pages.

\subsection{Structured stack LSTM decoder}
The decoder aims to decipher the motion from encoded features and output predicted poses frame by frame. Existing methods~\cite{Zhenguang2019Towards,Martinez2017On} utilized LSTM or GRU to achieve this goal, which regards different parts of the skeleton as equal important and breaks the structural principle of the skeleton. This inspires us to design a structured stack LSTM decoder with three layers. The first layer models overall information of motion from the encoder. Then, a new LSTM is used to predict the spine in the second layer, and another two LSTMs are utilized to obtain arms and legs in the last layers. At the first layer, the cell state input is set to $\hat{c}^0 =  \frac{1}{t-1}\sum\limits^{t-1}_{i=1}c^L_{i,\cdot}$, where $c^L_{i,\cdot} = [{c^L_{i,1}}^\mathrm{ T }, ..., {c^L_{i,K}}^\mathrm{ T }]^\mathrm{ T }$ and the hidden state input is $\hat{h}^0 = \frac{1}{t-1}\sum\limits^{t-1}_{i=1}h^L_{i,\cdot}$. For the second layer, the cell state and input state are $\hat{c}^1 = \frac{1}{t-1}\sum\limits^{t-1}_{i=1}c^L_{i,\cdot}$ and $\hat{h}^1 = \frac{1}{t}(\sum\limits^{t-1}_{i=1}h^L_{i,\cdot} + {g_t}^{L}_{\cdot})$, where ${g_t}^{L}_{\cdot} = [{{g_t}^{L}_{1}}^\mathrm{ T }, ..., {{g_t}^{L}_{K}}^\mathrm{ T }]^\mathrm{ T }$. At the rest of layers, the initial hidden state and cell state inputs are set to $\boldsymbol{0}$.

\subsection{Loss function}
Currently there are three loss functions commonly used during network training, i.e., calculating L2 loss on the Lie algebra vector directly or obtaining the locations of joints or bones by forward kinematics and computing their L2 loss. However, these functions neglect the kinematic relations among bones in chains and regard all the bones equally. To eliminate this problem, a new loss function~\cite{Zhenguang2019Towards} is presented by computing a weight for each element in the Lie algebra vector $\bm{\omega}$. The fact is that the prediction on a bone is much more important than that on its successive bones in a chain when doing forward kinematics. However, this function cannot quantize the accumulative effect of the bones in the back of the chain. Here we redefine the function, allowing it to estimate such effect, as follows:
\begin{equation}
\footnotesize
Loss(\boldsymbol{P}, \boldsymbol{\hat{P}}) = \frac{1}{T-t-1}\sum\limits^T_{i=t+1}\sum\limits_{z=1}^K\Theta(z)||\omega_{i,j}-\hat{\omega}_{i,j}||,
\end{equation}
\begin{equation}
\label{equ:equ1}
\footnotesize
\Theta(z) = \sum\limits^K_{j=z}(K+1-j)\ell_{i,j},
\end{equation}
where $\ell_{i,j}$ denotes the length of bone $j$ at frame $i$, and $\hat{\omega}_{i,j}$ refers to the predicted bone. $\Theta(z)$ indicates the weight of current bone $z$ by accumulating the lengths and locations of its successive bones. Consequently, a bone is given more penalty coefficient if it has longer subsequent bones.

\section{Experiment}
\subsection{Datasets}
Experiments are carried out on two benchmarks: Human3.6M~\cite{Catalin2014Human3} including 3.6 million accurate 3D human poses, on which we choose 15 activities of 7 subjects and down sample the FPS of a pose sequence to 25; Mouse dataset~\cite{Lie-X}, which records the motion of four mice in nine videos under lab condition. %Furthermore, we also focus the human's hand gesture prediction, which is an important application in the future for sign language generation. We use Kinect to record data and 6 people are participated. Each of them gesture circle, square, and triangle using their right hands in different videos. Like Human3.6M, the videos are down sampled to 25 FPS.

\subsection{Parameters}
\label{sec:sec1}
In our experiments, the length of hidden state in the encoder is set to 20 and 16 for H3.6m and muouse datasets, respectively. The recurrent step is 10 and batch size is 32. For short-term prediction, we randomly collect data samples with 60 consecutive frames from videos (i.e. $T=60$), which is the same as other comparative approaches~\cite{Tang2018Long,Zhenguang2019Towards}. The first 50 frames are used to feed the encoder and decoder, while the remaining 10 frames are left for the prediction. On the other hand, 50 frames are feeded into the network to predict 100 frames in long-term prediction. The Adam tool~\cite{Kingma2014Adam} is utilized to the optimize the network with its parameters $\beta_1$ and $\beta_2$ set to 0.9 and 0.999, respectively. Our model is implemented on Pytorch 1.0 and the model parameters are randomly initialized using Gaussian distribution.

\subsection{Baseline methods}
The prediction performance of our approach is compared against six established RNN-based methods: ERD and LSTM-3LR~\cite{Fragkiadaki2015Recurrent}, SRNN~\cite{Jain2016Structural}, MHU~\cite{Tang2018Long}, Res-GRU~\cite{Martinez2017On}, and HMR~\cite{Zhenguang2019Towards}. We rely on the code and pre-trained models of these methods to reproduce their work.
%For MHU, there is no code available so far, so we just utilized the results published on the paper.
These competing models are evaluated in two aspects: $quantitative$, which indicates angle errors in the short-term prediction, and $qualitative$, which considers feasible motion (dynamic and human like) in the long-term prediction. In particular, for quantitative evaluation we use the mean angle error (MAE) metric indicating the angle difference of two bones between the prediction and ground truth.
Also, we take agnostic zero-velocity~\cite{Martinez2017On} into consideration, which always regards the new predicted pose as the last observed pose. This method is significant to analyze the effectiveness of models in motion prediction as baseline.

\section{Results and discussion}

\subsection{Results on H3.6m dataset}
\begin{table*}
\centering
\scalebox{0.75}{
%\begin{tabular}{p{3cm}p{0.5cm}p{0.5cm}p{0.5cm}p{0.5cm}p{0.5cm}p{0.5cm}p{0.5cm}p{0.5cm}|p{0.5cm}p{0.5cm}p{0.5cm}p{0.5cm}p{0.5cm}p{0.5cm}p{0.5cm}p{0.5cm}}
\begin{tabular}{lrrrrrrrr|rrrrrrrr}
\toprule
\multirow{2}{*}{Methods} & \multicolumn{8}{c}{Greeting} & \multicolumn{8}{|c}{Walking} \\
\cmidrule{2-17}
                        & 80ms & 160ms & 320ms & 400ms & 560ms & 640ms & 720ms & 1000ms & 80ms & 160ms & 320ms & 400ms & 560ms & 640ms & 720ms & 1000ms \\
\midrule
ERD~\cite{Fragkiadaki2015Recurrent} & 1.152 & 1.321 & 1.582 & 1.692 & 1.912 & 1.922 & 1.943 & 2.010 & 1.061 & 1.123 & 1.221 & 1.263 & 1.311 & 1.342 & 1.412 & 1.512 \\
LSTM-3LR~\cite{Fragkiadaki2015Recurrent} & 0.922& 1.123 & 1.394 & 1.506 & 1.764 & 1.762 & 1.811 & 1.912 &0.882 & 0.952 & 1.018 & 1.053 & 1.102 & 1.120 & 1.142 & 1.211 \\
SRNN~\cite{Jain2016Structural} & 0.743 & 1.074 & 1.477 & 1.673 & 2.142 & 2.113 & 2.192 & 2.422 & 0.642 & 0.829 & 1.076 & 1.223 & 1.463 & 1.513 & 1.552 & 1.581 \\
Res-GRU~\cite{Martinez2017On} & 0.572 & 0.923 & 1.282 & 1.442 & 1.744 & 1.762 & 1.821 & 1.948 & 0.342 & 0.552 & 0.772 & 0.87 & 1.072 & 1.142 & 1.233 & 1.352\\
Zero-velocity~\cite{Martinez2017On} & 0.544 & 0.891 & 1.302 & 1.494 & 1.761 & 1.741 & 1.772 & 1.800 & 0.391 & 0.682 & 0.993 & 1.151 & 1.353 & 1.368 & 1.372 & 1.322\\
MHU~\cite{Tang2018Long} & \textbf{0.540} & 0.870 & 1.270 & 1.450 & 1.750 & 1.710 & 1.740 & 1.870 & 0.320 & 0.530 & 0.690 & 0.770 & 0.900 & 0.940 & 0.970 & 1.060\\
HMR~\cite{Zhenguang2019Towards} & 0.545 & 0.905 & 1.272 & 1.409 & 1.662 & 1.650 & 1.690 & 1.721 & 0.355 & 0.551 & 0.790 & 0.854 & 0.949 & 0.983 & 1.042 & 1.111\\
Ours(Remove ${g_t}_{j}^{l}$)  & 0.543 & 0.881 & 1.252 & 1.388 & 1.622 & 1.574 & 1.633 & 1.691 & 0.322 & 0.453 & 0.692 & 0.771 & 0.859 & 0.902 & 0.971 & 0.992\\
Ours(Remove ${g_s}_{i}^{l}$)  & 0.562 & 0.891 & 1.258 & 1.401 & 1.622 & 1.591 & 1.612 & 1.672 & 0.353 & 0.462 & 0.711 & 0.792 & 0.893 & 0.924 & 0.969 & 1.008\\
Ours(Replace LSTM)  & 0.544 & 0.881 & 1.242 & 1.383 & 1.601 & 1.578 & 1.631 & 1.682 & 0.321 & 0.452 & 0.704 & 0.772 & 0.869 & 0.911 & 0.962 & 1.001\\
Ours & 0.543 & \textbf{0.858} & \textbf{1.228} & \textbf{1.368} & \textbf{1.585} & \textbf{1.554} & \textbf{1.602} & \textbf{1.659} & \textbf{0.302} & \textbf{0.420} & \textbf{0.681} & \textbf{0.760} & \textbf{0.851} & \textbf{0.892} & \textbf{0.944} & \textbf{0.984}\\
\midrule
\multirow{2}{*}{Methods} & \multicolumn{8}{c|}{Posing} & \multicolumn{8}{|c}{Purchases} \\
\cmidrule{2-17}
                        & 80ms & 160ms & 320ms & 400ms & 560ms & 640ms & 720ms & 1000ms & 80ms & 160ms & 320ms & 400ms & 560ms & 640ms & 720ms & 1000ms \\
\midrule
ERD~\cite{Fragkiadaki2015Recurrent} & 1.353 & 1.413 & 1.691 & 1.863 & 2.064 & 2.115 & 2.183 & 2.568 & 1.162 & 1.300 & 1.492 & 1.522 & 1.812 & 1.856 & 1.849 & 2.340\\
LSTM-3LR~\cite{Fragkiadaki2015Recurrent} & 1.220 & 1.251 & 1.543 & 1.711 & 1.932 & 2.012 & 2.093 & 2.732 & 1.032 & 1.131 & 1.352 & 1.421 & 1.812 & 1.880 & 1.812 & 2.301\\
SRNN~\cite{Jain2016Structural} & 0.961 & 1.143 & 1.703 & 2.042 & 2.481 & 2.471 & 2.693 & 3.501  & 0.692 & 1.091 & 1.481 & 1.672 & 1.923 & 1.991 & 1.911 & 2.481 \\
Res-GRU~\cite{Martinez2017On} & 0.401 & 0.742 & 1.386 & 1.662 & 1.983 & 2.123 & 2.231 & 2.671 & 0.541 & 0.792 & 1.101 & 1.201 & 1.611 & 1.691 & 1.712 & 2.161 \\
Zero-velocity~\cite{Martinez2017On} & 0.281 & 0.572 & 1.132 & 1.372 & 1.812 & 2.143 & 2.227 & 2.780 & 0.621 & 0.881 & 1.192 & 1.269 & 1.643 & 1.681 & 1.624 & 2.451\\
MHU~\cite{Tang2018Long} & 0.330 & 0.640 & 1.220 & 1.470 & 1.820 & 2.110 & 2.170 & 2.510 & \_ & \_ & \_ & \_ & \_ & \_ & \_ & \_\\
HMR~\cite{Zhenguang2019Towards} & 0.239 &  0.509 & 1.058 & 1.310 & 1.636 & \textbf{1.802} & \textbf{1.942} & \textbf{2.485} & 0.514 & 0.776 & 1.053 & 1.150 & 1.602 & 1.665 & 1.610 & 2.110\\
Ours(Remove ${g_t}_{j}^{l}$)  & 0.230 & 0.511 & 1.062 & 1.333 & 1.632 & 1.859 & 2.018 & 2.602 & 0.529 & 0.790 & 1.071 & 1.122 & 1.541 & 1.569 & 1.521 & 2.121\\
Ours(Remove ${g_s}_{i}^{l}$)  & 0.252 & 0.515 & 1.072 & 1.333 & 1.648 & 1.872 & 2.012 & 2.611 & 0.528 & 0.780 & 1.052 & 1.140 & 1.551 & 1.582 & 1.544 & 2.133\\
Ours(Replace LSTM)  & 0.241 & 0.502 & 1.064& 1.321 & 1.628 & 1.868 & 2.012 & 2.601 & 0.524 & 0.801 & 1.054 & 1.124 & 1.523 & 1.558 & 1.511 & 2.131\\
Ours & \textbf{0.225} & \textbf{0.486} & \textbf{1.055} & \textbf{1.300} & \textbf{1.622} & 1.838 & 1.993 & 2.581 & \textbf{0.509} & \textbf{0.775} & \textbf{1.037} & \textbf{1.105} & \textbf{1.490} & \textbf{1.544} & \textbf{1.493} & \textbf{2.106}\\
\bottomrule
\end{tabular}}
\caption{The MAE comparisons on H3.6m dataset.}
\label{tab:benchmark1}
\end{table*}

The quantitative results of the complex activities, e.g. \emph{Greeting}, \emph{Walking}, \emph{Posing}, and \emph{Purchases}, are reported in Table~\ref{tab:benchmark1} (other activities are not shown due to page limitation, but similar results are provided in our home page). It can be observed that our model clearly outperforms the other models with a margin for both short-term and long-term prediction. This is mainly due to its abilities to take advantage of the rich spatio-temporal dependency information between bones in chains sperately.
Notably, \emph{Greeting} and \emph{Purchases} are more challenging than others because they contain more hand movements than leg and foot movements. Fortunately, our model can effectively encode hand movement and leg movement simultaneously. In addition, it is clear that the zero-velocity performance is better than that of ERD, LSTM-3LR and SRNN, which is consistent with the results in~\cite{Martinez2017On}. This might be that some activities only change their motions slightly. In such situation, zero-velocity can yield static poses continuously, but other competing models may suffer from the first frame discontinuity issue.
%Furthermore, the spatial and temporal global states not only can capture the movement of one bone over frame but also one pose'e integral information, which enhance the performance of our model.

For qualitative evaluation, we evaluate these models by visualizing the motions from two primary aspects: human-like and recognizable. Note that we only list 3 out of 11 activities here, and visualize 4 of the first 100 frames as short-term motion and the rest of 12 frames as long-term motion. We refer the interested readers to visit our project website for better visual effect.
For instance, all models perform well on short-term prediction on \emph{walking}, but for long-term prediction, LSTM-3LR, Res-GRU, and ERD converge to motionless state.
It is obvious that our model and HMR yield human-like and recognizable poses throughout the entire prediction window where the movement speed of our model is more close to the ground truth than HMR. This is mainly due to the global states being designed to encode integrated information at both spatial and temporal domain.
Besides, we found that \emph{walking} is relatively simple since it only contains repetitive movements of legs and arms. However, for a more complex activity \emph{eating}, which contains significant motions like feeding food to the mouth with hands, HMR only learns the foot movement but the hands are motionless pose. Other comparison models cannot obtain recognizable motion.
To further complicate the matter, in \emph{posing}, which contains motion features including standing still and doing several poses by hands, it is clear that only our model captures these features and repeatedly yields the motional and human like poses, as shown in Fig~\ref{fig:fig4}.

\begin{figure*}
\centering
\includegraphics[width=1\textwidth]{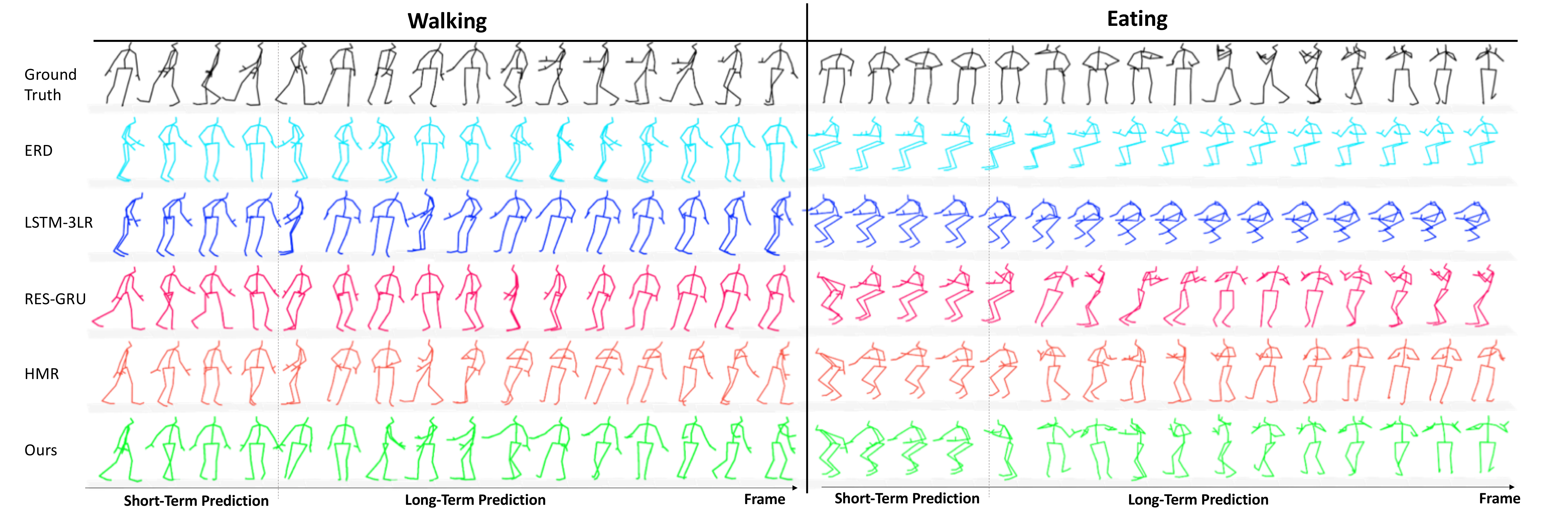}
\caption{Illustrations of qualitative comparisons on long-term motion prediction of \emph{walking} and \emph{eating} on H3.6m dataset.}
\label{fig:fig3}
\end{figure*}

\subsection{Results on mouse dataset}
\begin{figure}
\centering
\includegraphics[width=1.0\columnwidth]{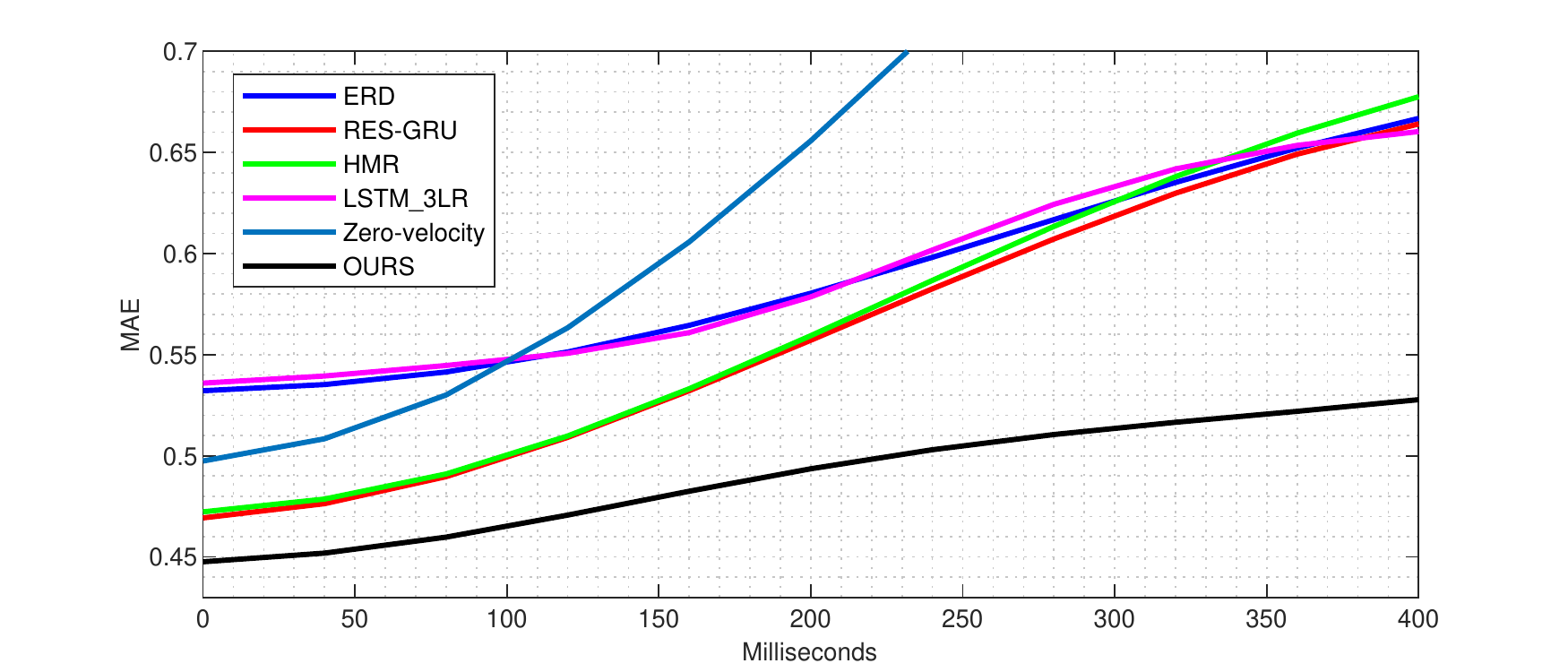}%{Fig5.eps}
\caption{Comparisons of short-term predictions on mouse dataset.}
\label{fig:fig5}
\end{figure}

Unlike human dataset, mouse dataset is more challenging due to its stochastic nature which causes difficulties to category its motion~\cite{Lie-X}.
Table~\ref{tab:benchmark2} depicts the comparison results with MAE. Our model outperformes other models on six out of eight frames.
We also found that zero-velocity only surpasses others at the $80$ms frame and falls behind with a notable margin on the remaining frames.
This is because the movement of mouse is faster and more random than the human. As suggested in Fig~\ref{fig:fig5}, our model outperforms others, which verifies the superiority and stability of our model.
\begin{table}
\centering
\scalebox{0.75}{
%\begin{tabular}{p{3cm}p{0.5cm}p{0.5cm}p{0.5cm}p{0.5cm}p{0.5cm}p{0.5cm}p{0.5cm}p{0.5cm}|p{0.5cm}p{0.5cm}p{0.5cm}p{0.5cm}p{0.5cm}p{0.5cm}p{0.5cm}p{0.5cm}}
\begin{tabular}{lrrrrrrrr}
\toprule
\multirow{2}{*}{Methods} & \multicolumn{8}{c}{Mouse}\\
\cmidrule{2-9}
                        & 80ms & 160ms & 320ms & 400ms & 560ms & 640ms & 720ms & 1000ms \\
\midrule
ERD~\cite{Fragkiadaki2015Recurrent} & 0.501 & 0.482 & 0.631 & 0.694 & 0.720 & 0.679 & 0.692 & 0.812 \\
LSTM-3LR~\cite{Fragkiadaki2015Recurrent} & 0.534 & 0.490 & 0.659 & 0.681 & 0.672 & 0.616 & 0.701 & 0.752 \\
Res-GRU~\cite{Martinez2017On} & 0.410 & 0.471 & 0.622 & 0.693 & 0.701 & 0.638 & 0.700 & \textbf{0.700}\\
Zero-velocity~\cite{Martinez2017On} & \textbf{0.400} & 0.531 & 0.732 & 0.951 & 1.028 & 0.941 & 1.069 & 1.131\\
HMR~\cite{Zhenguang2019Towards} & 0.420 & 0.441 & 0.642 & 0.711 & 0.728 & 0.709 & 0.731 & 0.720 \\
Ours & 0.410 & \textbf{0.428} & \textbf{0.533} & \textbf{0.521} & \textbf{0.570} & \textbf{0.501} & \textbf{0.668} & 0.721 \\
\bottomrule
\end{tabular}}
\caption{The MAE comparisons on mouse dataset.}
\label{tab:benchmark2}
\end{table}

\subsection{Ablation study}
\subsubsection{Loss function}
In this section, we evaluate the effectiveness of our proposed loss function by comparing it against L2 loss and the HMR loss~\cite{Zhenguang2019Towards} functions. We use H3.6m dataset for this study with the same parameter settings depicted in section~\ref{sec:sec1}. Table~\ref{tab:benchmark3} reports the comparison results of average MAEs on all the activities. Our loss function completely outperforms L2 loss and HMR loss. This is because our loss function considers the error accumulation effect when estimating the root bones. It not only remains the anatomical restricts of chains, but also provides a bone with a quantized weight in terms of its location and length in a pose.
\begin{table}
\centering
\scalebox{0.75}{
%\begin{tabular}{p{3cm}p{0.5cm}p{0.5cm}p{0.5cm}p{0.5cm}p{0.5cm}p{0.5cm}p{0.5cm}p{0.5cm}|p{0.5cm}p{0.5cm}p{0.5cm}p{0.5cm}p{0.5cm}p{0.5cm}p{0.5cm}p{0.5cm}}
\begin{tabular}{lrrrrrrrr}
\toprule
\multirow{2}{*}{Loss} & \multicolumn{8}{c}{H3.6m}\\
\cmidrule{2-9}
                        & 80ms & 160ms & 320ms & 400ms & 560ms & 640ms & 720ms & 1000ms \\
\midrule
L2 loss & 0.362 & 0.608 & 0.971 & 1.101 & 1.303 & 1.396 & 1.472 & 1.803 \\
HMR loss~\cite{Zhenguang2019Towards} & 0.340 & 0.600 & 0.950 & 1.060 & 1.290 & 1.370 & 1.450 & 1.770 \\
Our loss  & \textbf{0.331} & \textbf{0.579} & \textbf{0.931} & \textbf{1.058} & \textbf{1.283} & \textbf{1.359} & \textbf{1.441} & \textbf{1.750}\\
\bottomrule
\end{tabular}}
\caption{The MAE performance of our model on different loss functions.}
\label{tab:benchmark3}
\end{table}

\subsubsection{Component effectiveness}
We separately evaluate the effects of different components in our network by removing modules or replacing them with conventional methods.
They are evaluated by testing for two types of investigations that are common with neural network models: encoder component effects (i.e. remove temporal states ${g_t}_{j}^{l}$ and spatial states ${g_s}_{i}^{l}$, respectively) and decoder component effects (i.e. replace our structured stack LSTM decoder with a na\"{i}ve LSTM of two layers). Table~\ref{tab:benchmark1} shows that changing the components may lead to negative effects on the performance of our model. It is clear that when removing spatial states ${g_s}_{i}^{l}$, the prediction performance drops faster than that changing other components. This might be due to the high-level encoding of fine-garined spatial information in our model. Besides, when using the na\"{i}ve LSTM decoder, the model gives worse performance than that using our structured stack LSTM decoder, which indicates that our model is more effective to predict spine, arms, and legs gradually than obtaining them at the same time.

\section{Conclusion}
In this paper, we present a spatio-temporal hierarchical recurrent network, where the hierarchical model is incorporated to simultaneously capture the inherit spatial and temporal varieties of motions. It is more efficient and flexible than existing methods on both short-term and long-term motion predictions. As for future work, we will explore the applications of our model on raw image videos, and we will consider predicting multiple motions with probabilities and will instead learn a network generating future motions under uncertainty.

\ack This work was supported by grants from the National Major Science and Technology Projects of China (grant nos. 2018AAA0100703, 2018AAA0100700), the National Natural Science Foundation of China (grant no. 61977012), the Chongqing Provincial Human Resource and Social Security Department (grant no. cx2017092), the Central Universities in China (grant nos. 2019CDJGFDSJ001).

\bibliography{ecai}
\end{document}